\pdfoutput=1

\documentclass[11pt]{article}

\usepackage{emnlp2021}

\usepackage{times}
\usepackage{latexsym}
\usepackage{graphicx}

\usepackage[T1]{fontenc}

\usepackage[utf8]{inputenc}
\usepackage{multirow}
\usepackage{bm}
\usepackage{tabularx}
\usepackage{microtype}

\title{Teach Me What to Say and I Will Learn What to Pick: \\ Unsupervised Knowledge Selection Through Response Generation with Pretrained Generative Models}

\author{Ehsan Lotfi, Maxime De Bruyn, Jeska Buhmann, Walter Daelemans \\
        CLiPS Research Center \\ University of Antwerp, Belgium \\
        \texttt{firstname.lastname@uantwerpen.be}
}

\begin{document}
\maketitle
\begin{abstract}
Knowledge Grounded Conversation Models (KGCM) are usually based on a selection/retrieval module and a generation module, trained separately or simultaneously, with or without having access to a `gold' knowledge option. With the introduction of large pre-trained generative models, the selection and generation part have become more and more entangled, shifting the focus towards enhancing knowledge incorporation (from multiple sources) instead of trying to pick the best knowledge option. These approaches however depend on knowledge labels and/or a separate dense retriever for their best performance.
In this work we study the unsupervised selection abilities of pre-trained generative models (e.g. BART) and show that by adding a score-and-aggregate module between encoder and decoder, they are capable of learning to pick the proper knowledge through minimising the language modelling loss (i.e. without having access to knowledge labels). Trained as such, our model -  K-Mine - shows competitive selection and generation performance against models that benefit from knowledge labels and/or separate dense retriever.      

\end{abstract}

\section{Introduction}
The ability to properly ground conversations in structured and unstructured data, has become an increasingly important feature in designing conversational agents. By generating more informative and specific responses, such models can establish human-machine interactions that are more engaging and less prone to producing bland and common responses.
The task of modelling knowledge-grounded conversations is traditionally decomposed into two sub-tasks: 1) knowledge selection (\textbf{KS}), i.e. picking the proper knowledge piece(s) from a provided pool based on dialogue history, and 2) response generation (\textbf{RG}), i.e. producing a response to the user's utterance conditioned on both dialogue history and selected knowledge piece(s). Therefore and because of this sequential dependency, the generation performance is directly affected by model's selection/retrieval ability and the way this knowledge is being incorporated in the generation process.

Early examples of knowledge grounded conversation models mainly tried to diffuse the external knowledge as an extra hidden state into the decoder part of a recurrent seq-to-seq architecture  \cite{liu-etal-2018-knowledge, ghazvininejad2018knowledgegrounded}. With the release of large knowledge grounded conversational datasets like Wizard of Wikipedia \cite{dinan2019wizard}, Topical-chat \cite{Gopalakrishnan2019} and Holl-E \cite{moghe2018exploiting}, the field witnessed numerous studies aimed to best coordinate the KS and RG sub-tasks to improve the overall performance of models. As an early standard baseline \citet{dinan2019wizard} proposed variations of Transformer MemNet, a generative model trained to do KS and RG using a memory network for selecting the most relevant knowledge piece.

\begin{figure*}[h]
    \centering
    \includegraphics[width=0.65\textwidth]{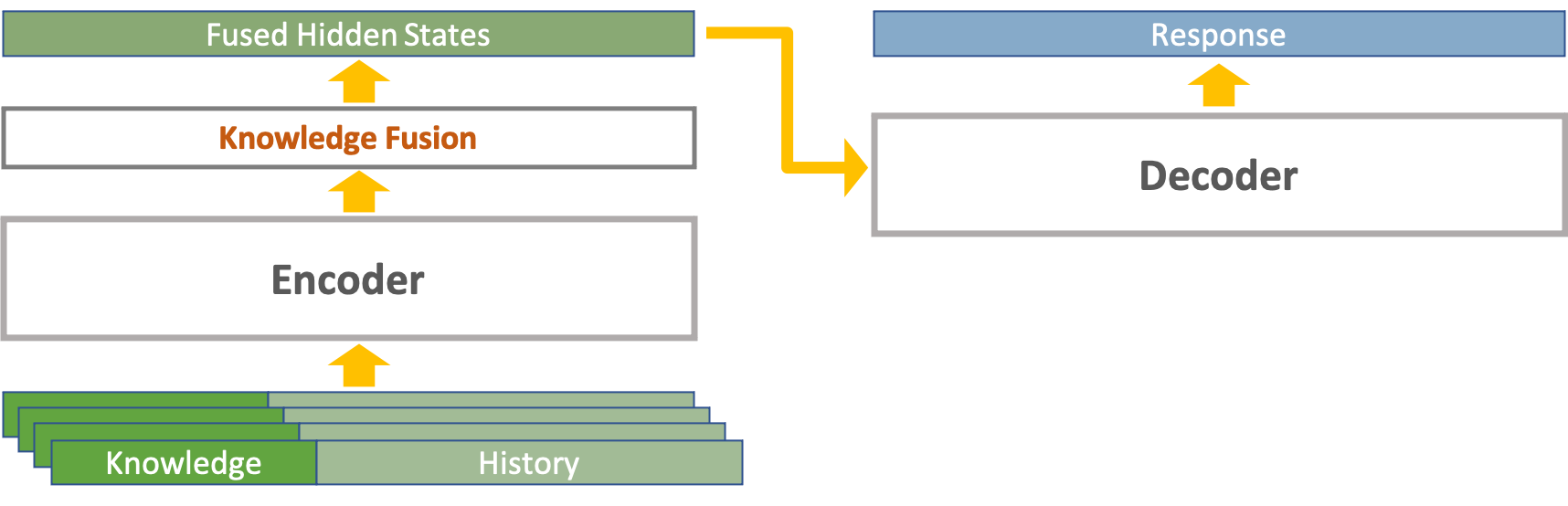}
    \caption{General overview of our model based on a pretrained encoder-decoder like BART}
    \label{fig:model_gen}
\end{figure*}

\label{model-types}
Attempts to improve on these benchmarks can mostly be divided into two categories, based on their point of focus. \textbf{Selection oriented methods} focus on enhancing the KS task, usually by introducing additional learning signals like the prior-posterior discrepancy \cite{lian2019learning, chen-etal-2020-bridging} or long-term structural traits of conversations like flow and initiative changes \cite{kim2020sequential, 10.1145/3397271.3401097, zhan-etal-2021-augmenting, Meng2021InitiativeAwareSL, zheng2020differenceaware}. \textbf{Generation oriented methods} on the other hand, try to mitigate the selection bottleneck by employing more powerful methods to incorporate knowledge in the generation process, thus reformulating the KS problem as an adaptive fine-grained selection to be dealt with in decoding \cite{10.1145/3357384.3357889}. This was especially encouraged with the introduction of large pretrained generative models like GPT2 \cite{Radford2019LanguageMA}, BART \cite{lewis2019bart} and T5 \cite{raffel2020exploring} which allow leveraging their ability in conditional text generation \cite{zhao-etal-2020-knowledge, Bruyn2020BARTFK}. More recently RAG \cite{lewis2021retrievalaugmented} and FiD \cite{izacard2021leveraging} models have been proposed (primarily in the QA context) to ease the computational costs and limitations of these big models, especially if the supporting passage(s) needs to be retrieved from a huge unstructured corpus. Since these models integrate the KS and RG tasks, they do not need labeled knowledge for training, although a knowledge pool of manageable size is provided often via a parametric or non-parametric retrieval module.    

In this study we propose  K-Mine (\underline{K}nowledge \underline{M}ixing \underline{in} \underline{e}ncoder); a model that bridges between the two aforementioned paradigms by doing unsupervised knowledge selection with and within pretrained generative models (e.g. BART). Using a simple score-and-aggregate module between the encoder and decoder of such a model,  K-Mine learns to (soft-) select the most relevant passage without needing knowledge labels or a separate retriever, while maintaining the generative skills of the original pretrained model. Our experiments show very competitive performances on two knowledge grounded conversation datasets and state of the art results in integrated unsupervised knowledge selection.     

\begin{figure*}[h]
    \centering
    \includegraphics[width=0.85\textwidth]{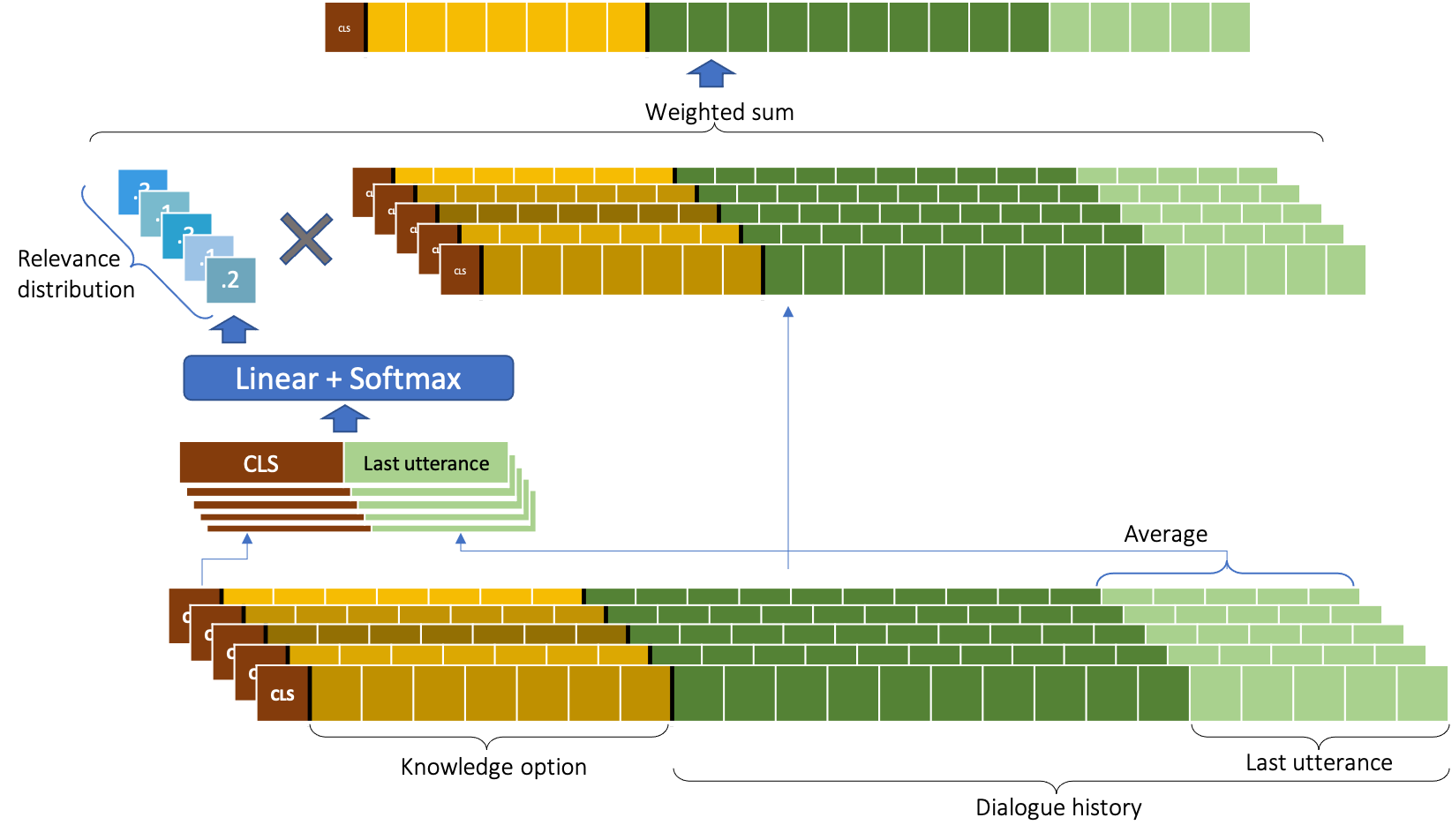}
    \caption{Inside the knowledge fusion module: Inputs are the encoded knowledge-history pairs (same history with different (in this case 5) knowledge options) from which the [CLS] embedding and the average last utterance embedding are concatenated and fed to a linear layer to calculate relevance scores. The scores are converted to a distribution via a normalizing function like Softmax. Finally the normalized scores are used to produce a weighted sum of the inputs which will be passed to the decoder as the encoder output.}
    \label{fig:km}
\end{figure*}

\section{Related Work}
Like most NLP tasks, knowledge grounded conversation has been significantly influenced by the introduction of large pretrained language models, which have helped generative models beat retrieval models in both automatic and human evaluations \cite{roller2020opendomain}. Thanks to their language modeling skills, these models have shown the ability to use the provided context (e.g. history, knowledge, persona, etc.) on a per-demand basis, thus alleviating the knowledge selection bottleneck. Adapting them to the specifics of the problem, usually requires modifications in such a way that would still allow for leveraging model's pretrained skills. \citet{zhao-etal-2020-knowledge} employed reinforcement learning to jointly optimize KS and RG with unlabeled dialogues in a BERT-GPT2 based architecture. \citet{Bruyn2020BARTFK} opted for BART by modifying the encoder to do the supervised KS task and showed that providing the decoder with more knowledge pieces (top-k instead of 1), leads to a lower RG perplexity. \citet{izacard2021leveraging} proposed Fusion in Decoder (FiD) which passes a selection of individually encoded question-knowledge pairs as a single concatenated sequence to decoder, hence reducing the computational cost of self-attention over all knowledge options at once. RAG \cite{lewis2021retrievalaugmented} is another (more general) framework to incorporate knowledge in text generation which allows the (pretrained) decoder to choose content from top-k retrieved knowledge pieces via token-wise or sequence-wise marginalization. In principle FiD and RAG replace the KS step with a pool retrieval task that provides the pretrained model with multiple (top-k) relevant passages to attend to (FiD) or marginalize over (RAG) during generation. In particular RAG benefits from a DPR retriever which is updated (only the query encoder part) during training through the back-propagation of generation error. Recently \citet{shuster2021retrieval} adopted FiD and RAG (originally introduced mainly for QA tasks) for knowledge grounded conversation and tried to improve the performance with a variety of general and task-inspired modifications, e.g. using poly encoders \cite{humeau2020polyencoders} or extending the marginalized decoding idea to dialog turns.

Our work has similarities with both FiD and RAG in the sense of learning knowledge grounded generation with pretrained models and without the need to have labeled knowledge. It is however different in some key aspects. The retrieval/selection part in  K-Mine is truly and completely integrated inside the pretrained model, so unlike RAG, it does not require a separate parametric retriever, and unlike FiD, it is not totally disentangled from the retriever. Moreover, unlike both FiD and RAG,  K-Mine aggregates over encoded knowledge options before passing them to the decoder. We will discuss the advantages and disadvantages of these choices at the end of the paper.

\section{Methodology}
\subsection{Problem Definition and Formalization}
In general the question of knowledge grounded conversation modelling is defined over dialog and knowledge datasets $\mathcal{D}_{d} = \{(C_{i}, r_{i})\}_{i=1}^{N}$ and $\mathcal{D}_{k} = \{(k_{j})\}_{j=1}^{M}$ where $\forall i \in \{1,...,N\}$, $C_{i}$ and $r_{i}$ represent context and response for a specific dialog turn, and $\forall j \in \{1,...,M\}$, $k_{j}$ is a knowledge piece (e.g. a sentence or paragraph). $\mathcal{D}_{d}$ and $\mathcal{D}_{k}$ usually are connected through a retrieval function: $$f_{ret} : \mathcal{D}_{d} \rightarrow \mathcal{D}_{k}^{m} \; ; \; m \in \{0,...,M\} \: , m << M$$
$f_{ret}$ can be part of the trained model or a non-parametric module which does a preliminary filtering by narrowing down the knowledge options from $M$ to $m$, based on some similarity metric. In most knowledge grounded conversation datasets, $\mathcal{D}_{d}$ and $\mathcal{D}_{k}$ are provided as parallel, which allows for a simpler formalization over $\mathcal{D} = \{(C_{i}, K_{i}, r_{i})\}_{i=1}^{N}$, where $K_{i} \in \mathcal{D}_{k}^{m} ; \; m \in \{0,...,M\}$ is the narrowed down subset of the original $\mathcal{D}_{k}$, and often includes a `gold truth', i.e. the knowledge piece which has been picked by the (human) participant during data curating. We consider the model to be `Fully-supervised' if this gold truth is used in training. Otherwise, it will be referred to as `RG-supervised'.

\begin{table*}
\centering
\begin{tabular}{ {l}*{8}{c} }
\hline
\textbf{Dataset} & \multicolumn{2}{c}{\textbf{Train}} & \multicolumn{2}{c}{\textbf{Valid}} & \multicolumn{2}{c}{\textbf{Test}} & \textbf{\#Kn}   \\
\hline
 & \textbf{All} & \textbf{w/ Kn}&\textbf{All} & \textbf{w/ Kn}&\textbf{All} & \textbf{w/ Kn}&\\
\hline
Wizard of Wikipedia & 82965 & 77332 & 8814 & 8270 & 4336/4370 & 4073/4119 &63 \\
\hline
HOLL-E & 36584 & 34632 & 4654 & 4399 & 4602 & 4339 & 58 \\
\hline
\end{tabular}
\caption{\label{datasets}
Overview of datasets used in the study. Numbers are for turns with access to knowledge and \textit{w/ Kn} refers to the number of such turns for which a knowledge option has been chosen to generate the response. WoW test set is divided into seen/unseen subsets and \textit{\#Kn} is the (average) number of knowledge options provided for each turn.}
\end{table*} 

\subsection{Approach and model }
Figure \ref{fig:model_gen} shows a general overview of our model which is built by adding a `knowledge fusion' module between the encoder and decoder of a pretrained generative model like BART. The encoder receives the input sequence as the concatenation of the $[CLS]$ token, a knowledge option and the dialog history which includes the last $u$ utterances, each pre-pended by special identifier tokens $<\!user\!>$ or $<\!agent\!>$ based on the speaker. One learning sample contains $m$ such sequences (padded to the same length) with different knowledge options and same history, which allows the encoder to create $m$ contextualized encodings for each knowledge-history pair. These encodings then are passed to the knowledge manager module which fuses them and creates a single sequence of hidden states to be fed to the decoder as the final encoder output. In order to make the manipulation and modification of hidden states easier in the knowledge fusion module, we pad-truncate knowledge options to the same length, decided by the dataset statistics.

Figure \ref{fig:km} shows a detailed overview of how the fusion is done. The module receives the encoded knowledge-history pairs (created by the pretrained encoder), from which the [CLS] token embedding and the average embedding of the last utterance are concatenated and fed to a linear layer to calculate relevance scores for each pair. The scores then are converted into a distribution by a normalizing function (e.g. Softmax). Finally the normalized scores are used to produce a weighted sum of the inputs which will be passed to the pretrained decoder as the encoder hidden states. Empirically the fused output can be written as: $$h_{j}^{fus}=\sum_{i=1}^{m} \alpha_{i}H_{ij}^{enc}$$  $$\alpha_{i}=f([CLS_{i}; mean(LU_{i})])$$ where $m$ is the number of knowledge options for each sample, $H_{i}^{enc}$ is the encoded hidden states for the $i_{th}$ knowledge-history pair, $LU$ is the last utterance and $f$ is the $Softmax(Linear)$ operator. The training is done by minimizing the usual NLL loss over the generated response in decoder. The knowledge fusion module can be also supervised by introducing a BCE loss with respect to the gold truth (when available). So in general: 
\begin{equation} \label{eq:1}
\mathcal{L} = (1-\lambda)\mathcal{L}_{RG} \:+ \lambda\mathcal{L}_{KS}
\end{equation}
although the RG-supervised case ($\lambda=0$) is of more interest to us.

\section{Experiments}
\subsection{Data}
We study our model on 2 publicly available datasets for knowledge grounded conversation (KGC): \\
\textbf{Wizard of Wikipedia (WoW)}\cite{dinan2019wizard} is a widely used dataset for open-domain KGC created by crowd-sourcing dialogues between an \textit{apprentice} and a \textit{wizard} who has access to a retrieved pool of Wikipedia passages which he/she can use in conversing with the apprentice. WoW consists of 22311 conversations (split into train, valid and test) over 1365 general topics. The validation and test set are further split into \textit{seen} and \textit{unseen} versions where the latter contains dialogues with new topics not discussed in the training data, for out-of-distribution topic evaluations. The knowledge pool size varies and on average each wizard turn is provided with \texttildelow 63 Wikipedia passages, although not all wizard turns make use of these options in generating the response\footnote{We use the pre-processed dataset provided by \citet{10.1145/3357384.3357889} in \url{https://github.com/ChuanMeng/DukeNet/}.}.
\\\\
\textbf{HOLL-E} \cite{moghe2018exploiting} is another KGC dataset that contains 7228, 930 and 913 dialogues for training, validation and test. Each conversation is about a specific movie and both parties have access to a document which contains the plot and a fact table besides a selection of viewer comments and reviews. The original dataset provides a list of spans in the document as knowledge options and indicates the one (if any) that has been used to generate the response. We use the processed version provided by \citet{kim2020sequential} which changed it to the WoW format by redefining the spans as complete sentences.   
\\

Table \ref{datasets} summarizes important information for these datasets. To have a more detailed evaluation, we do the experiments under two data settings: \textbf{w/Kn}; i.e. only including turns which use knowledge and \textbf{All}; i.e. including all turns.

\subsection{Evaluation metrics}
Following the related literature, we employ commonly used automatic metrics; \textbf{R@1} for knowledge selection, and  unigram \textbf{F1}, \textbf{ROUGE} and \textbf{PPL} (perplexity) for response generation. When comparing with variations (Table \ref{wow-abalation}) we also use \textbf{KF1} or Knowledge-F1 introduced by \citet{shuster2021retrieval} which measures the unigram word overlap between the model’s generation and the ground-truth knowledge. Whereas F1 can be seen as measuring conversational ability, KF1 attempts to capture whether a model is speaking knowledgeably by using relevant knowledge as judged by humans. This provides an easy way to distinguish between general language modeling skills and knowledge incorporation.

\subsection{Architecture and baselines}
In theory  K-Mine can be implemented using any pretrained encoder-decoder model including the two most commonly used ones, BART\cite{lewis2019bart} and T5 \cite{raffel2020exploring}. Exploring both options, BART turned out to yield better results so we opted for this model. We compared K-Mine with the following models: 
\\
\textbf{TMemNet} \cite{dinan2019wizard}: Combines a transformer (not pretrained) with an external memory network to select the knowledge. TMemNet+BERT, uses BERT as encoder. \\ 
\textbf{DukeNet} \cite{10.1145/3357384.3357889}: Explicitly models knowledge tracking and knowledge shifting as dual tasks to address the prior-posterior gap. It uses a BERT encoder.
\\
\textbf{MIKe} \cite{Meng2021InitiativeAwareSL}: Further improves KS by explicitly distinguishing between user-initiative and system-initiative knowledge selection. It uses a BERT encoder. 
\\
\textbf{BFKGC} \cite{Bruyn2020BARTFK}: Uses a BART-based model to do both KS and RG in a fully supervised manner (shared encoder). 
\\
\textbf{FiD-RAG} \cite{shuster2021retrieval}: Augments FiD by using a separate DPR-based retriever trained with RAG which results in state of the art performance on WoW.  

In addition, we also include a few specialized baselines/variations to better assess the performance. These include:
\\
\textbf{ K-Mine-mean}:  Instead of the weighted fusion, the decoder receives the average of knowledge-context encodings.
\\
\textbf{ K-Mine-max}: Instead of the weighted fusion, the decoder receives the argmax; i.e. the knowledge-context encoding with the highest relevance score. 
\\
\textbf{K-Mine-max-full}: Fully supervised K-Mine-max; i.e. with access to knowledge labels.
\\
\textbf{K-Mine-full}: Fully supervised K-Mine; i.e. with access to knowledge labels ($\lambda=.5$ in Equation \ref{eq:1}). 
\\
\textbf{KS-RoBERTa}: A RoBERTa model trained (only) on the knowledge selection task as a ranking problem with KS labels.

\subsection{Implementation details}
We use HuggingFace's Transformers library \cite{wolf2020huggingfaces} to implement our models. Training was done with an effective batch size of 64 and learning rates of 2e-5 and 5e-4 for the pretrained and raw parts respectively, with linear decay applied to both. For WoW dataset, we considered the last 3 utterances as the history, whereas for HOLL-E we just kept the last one since utterances mostly stand alone in this dataset. Passages were truncated or padded to the fixed length of 32 tokens before being concatenated with the history tokens, so that the weighted summation would not perturb the knowledge-history division in the input sequence.

\begin{table*}
\centering
\begin{tabular}{ l|| c| c c c || c| c c c  } 
\hline
 & \multicolumn{4}{c}{\textbf{Test Seen}} & \multicolumn{4}{c}{\textbf{Test Unseen}}  \\
\hline
\textbf{Model}& \textbf{R@1} & \textbf{PPL} & \textbf{F1} & {\small \textbf{ROUGE-L}}  & \textbf{R@1} & \textbf{PPL} & \textbf{F1} & {\small \textbf{ROUGE-L}}  \\
\hline
\hline
TMemNet (E2E)\textsuperscript{\textdagger}   &21.6 & 63.5 & 16.9 & 16.8 & 12.1 & 97.3 & 14.4 & 15.4 \\
\hline
TMemNet+BERT (E2E)\textsuperscript{\textdagger}    & 23.9 & 53.2 & 17.7 & 17.0 & 16.3 & 137.8 & 13.6 & 15.6  \\
\hline
DukeNet\textsuperscript{\textdagger}    & 26.4 & - & - & 18.5 & 19.6 & - & - & 17.0 \\ 
\hline
MIKe \textsuperscript{\textdagger}   & 28.4 & - & - & 18.8 & 21.5 & - & - & 17.4 \\ 
\hline
BFKGC\textsuperscript{\textdagger} {\small (BART-large)}   & 26.0 & 12.2 & 20.1 & - & 19.9 & 14.9 & 19.3 & - \\
\hline
FID-RAG\textsuperscript{\textdaggerdbl{}} {\small (BART-large)}    & {29.3}\textsuperscript{\textasteriskcentered} & 10.5 & 23.2 & 24.2 & 26.9\textsuperscript{\textasteriskcentered} & 10.7 & 23.2 & 24.4  \\
\hline  
\hline
\hline
 K-Mine  {\small (BART-base)} &  27.9 & 16.3 & 20.9 & 19.6  & 27.0 & 20.3 & 20.1 & 19.2 \\ 
\hline
 K-Mine -w/Kn {\small (BART-base)}&  29.2 & 16.1 & 21.4 & 19.9  & 28.4 & 20.3 & 20.3 & 19.4 \\ \hline
 K-Mine {\small (BART-large)} &  28.3 & 13.2 & 21.8 & 20.1  & 28.4 & 16.4 & 21.1 & 19.7 \\ 
\hline
 K-Mine -w/Kn {\small (BART-large)}&  30.4 & 13.1 & 22.2 & 20.1  & 30.8 & 16.5 & 21.5 & 20.0 \\

\end{tabular}
\caption{\label{wow-results-all} Results on WoW test sets for K-Mine and baselines. R@1 is the KS accuracy and the other metrics assess the response generation performance. Models with \textdagger{} next to their names benefit from KS labels in training and \textdaggerdbl{} identifies pretrained models which use a separate retriever. Results with \textasteriskcentered{} are for retrieval from 21M 100-word passages in Wikipedia so they are not directly comparable. K-Mine-w/Kn has been trained and tested on the w/Kn subsets; i.e. turns which have used knowledge (see Table \ref{datasets} for statistics) 
}

\vspace{2em}

\centering
\begin{tabular}{ l|| c| c c } 
\hline
\hline
\textbf{Model}& \textbf{R@1} & {\small \textbf{ROUGE-1}} & {\small \textbf{ROUGE-L}}   \\
\hline
\hline
TMemNet+BERT\textsuperscript{\textdagger}  & 28.4 & 31.6 & 25.9   \\
\hline
DukeNet\textsuperscript{\textdagger}  &  30.0 & 36.5 & 31.5  \\ 
\hline
MIKe\textsuperscript{\textdagger}   &  31.9 & 37.8 & 32.8 \\  
\hline
\hline
 K-Mine {\small (BART-base)} &  28.7 & 36.1 & 32.6   \\ 
\hline
 K-Mine -w/Kn {\small (BART-base)} & 30.7  & 37.3 & 33.8  \\ 
\hline
 K-Mine {\small (BART-large)} &  31.7 & 38.5 & 35.1   \\ 
\hline
 K-Mine -w/Kn {\small (BART-large)} &  32.8 & 39.3 & 36.0  \\ 
\end{tabular}
\caption{\label{holle-results-all} Results on HOLL-E test set (single reference) for  K-Mine and baselines. R@1 is the KS accuracy and the other metrics assess the response generation performance. Models with \textdagger{} next to their names, benefit from KS labels in training. K-Mine-w/Kn has been trained and tested on the w/Kn subsets; i.e. turns which have used knowledge (see Table \ref{datasets} for statistics)
}

\vspace{2em}

\centering
\begin{tabular}{ l|c|| c| c c c c || c| c c c c } 
\hline
 & &\multicolumn{5}{c}{\textbf{Test Seen}} & \multicolumn{5}{c}{\textbf{Test Unseen}}  \\
\hline
\textbf{Model} & \textbf{Data} & \textbf{R@1} & \textbf{PPL} & \textbf{F1} & \textbf{R-L}  & \textbf{KF1} &\textbf{R@1} & \textbf{PPL} & \textbf{F1} & \textbf{R-L} & \textbf{KF1} \\
\hline
\hline
K-Mine  & All & 27.9 & 16.3 & 20.9 & 19.6& 18.2  & 27.0 & 20.3 & 20.1 & 19.2& 17.5 \\ 
\hline
K-Mine & w/Kn  & 29.2 & 16.1 & 21.4 & 19.9& 19.1  & 28.4 & 20.3 & 20.3 & 19.4& 18.6 \\ 
\hline
K-Mine-mean & All &   - & 19.7 & 18.7 & 18.2 &14.4 & - &26.6&17.0&17.2&12.3 \\ 
\hline
K-Mine-mean & w/Kn & -  & 19.5 & 18.9& 18.2 & 15.4 & - & 27.0 & 17.2 & 17.3&13.1 \\ 
\hline
K-Mine-max & All & 3.2  & 18.5 & 18.5 & 18.2 & 14.1& 2.6 & 24.2 &16.9 &17.2 &12.1 \\ 
\hline
K-Mine-max & w/Kn & 1.2   & 18.5& 18.8& 18.3& 15.0& 0.9& 24.7&17.3&17.6&12.9 \\ 
\hline
K-Mine-max-full\textsuperscript{\textdagger} & All & 29.7   &17.2&20.8&19.4&19.8& 27.2& 20.8&19.9&18.9&18.7 \\ 
\hline
K-Mine-max-full\textsuperscript{\textdagger} & w/Kn &31.2 & 17.0 & 21.4 &19.8& 22.2& 29.0&20.7&20.5&19.4&21.1 \\ 
\hline
K-Mine-full\textsuperscript{\textdagger} & All & 29.7   &17.8&20.6&19.2&18.2& 28.3& 21.0&20.0&19.0&17.2 \\ 
\hline
K-Mine-full\textsuperscript{\textdagger} & w/Kn &30.9 & 17.6 & 21.1 &19.5& 19.9& 29.5&20.9&20.1&19.1&19.7 \\ 
\hline
\hline
KS-RoBERTa\textsuperscript{\textdagger} & All & 25.9    &-&-&-&-& 25.5 &-&-&-&- \\ 
\hline
KS-RoBERTa\textsuperscript{\textdagger}  & w/Kn & 28.3   &-&-&-&-&27.5&-&-&-&- \\

\end{tabular}
\caption{\label{wow-abalation} Results on WoW test sets for K-Mine and some extreme variations/baselines. R@1 is the KS accuracy and the other metrics assess the response generation performance (R-L = ROUGE-L). Models with \textdagger{} next to their names, benefit from KS labels in training. All models use the base version of the pretrained transformer. 
}
\end{table*}

\section{Results and discussion}
Tables \ref{wow-results-all} and \ref{holle-results-all} show the performance of  K-Mine in knowledge selection and response generation for the WoW (Seen and Unseen) and HOLL-E test sets against baselines. As one can see, K-Mine shows very competitive results, especially in knowledge selection accuracy (R@1) although it does not benefit from knowledge labels in training, or a separate retrieval module. Adding KS supervision (K-Mine-full) enhances the knowledge selection performance by\texttildelow 3\% although it negatively affects the RG performance. Regarding the data, the KS performance boosts by more than 2\% when only the `w/Kn' turns (turns that use knowledge to generate response) are used for training and testing, which shows that in the absence of explicit (KS) labels, the model prefers to select something than nothing\footnote{This might also depend on the way the `empty' or `no-knowledge' option is being represented. However in our experiments, using various choices including the original `no\_passages\_used' string, PAD tokens and the pool average, showed no difference. We also tried using certainty thresholds and gating mechanisms to link the 'no-knowledge' case with the relevance distribution but these too proved ineffective.}.

Table \ref{wow-abalation} shows the performance of standard K-Mine models against some  variations. Evidently the performance drops significantly when instead of the weighted mix, only the highest scored knowledge-history pair is passed to decoder (K-Mine-max vs. K-Mine), which shows that the full aggregated gradient is essential for the unsupervised KS learning. Adding KS supervision to K-Mine-max (K-Mine-max-full), boosts the performance in both KS and RG tasks, surpassing the standard K-Mine model. Here the KF1 metric is of special importance as it shows that although the fully supervised version (K-Mine-max-full) does slightly worse than K-Mine in terms of conversational ability and language modeling (F1, R-L and PPL), it generates significantly more knowledgeable responses, which can be attributed to passing a less noisy representation to decoder (as confirmed by the results from the K-Mine-full model). Finally, comparing the two fully supervised models (K-Mine-max-full and K-Mine-full) shows that in the presence of knowledge labels, fusion actually hurts the model's performance. \\  
Another interesting result is the higher discrepancy between the seen/unseen KS performance (R@1) in the presence of supervision (1.4-2.5 vs. 0.9) which indicates a lower ability to generalize when the task has been learned via explicit signals. Finally the inferior performance of the pure selection model (KS-RoBERTa) compared to K-Mine, K-Mine-max-full and K-Mine-full, reflects the broadly studied prior-posterior gap in KGCMs; i.e. the model can benefit from (or even exclusively rely on) responses for selecting the relevant knowledge\footnote{Comparing RoBERTa and BART in this manner is not trivial but it is also not unsound considering that the number of parameters are close (139M vs. 125M) and the selection task in K-Mine is `mainly' done via the BART encoder.}.

\begin{figure}[t]
    \centering
    \includegraphics[width=0.5\textwidth]{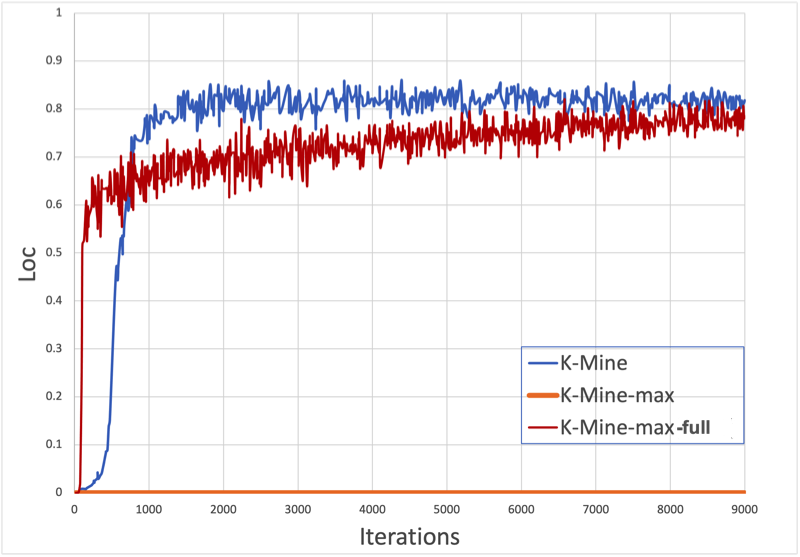}
    \caption{Evolution of the Localization metric during training for K-Mine and 2 variations. K-Mine-max is the (almost) flat line in the bottom.}
    \label{fig:loc}
\end{figure}

To have a better understanding of the way knowledge selection happens under the hood, we introduce a localization metric to measure the level by which the knowledge distribution vector $\bm{p}$ deviates from the uniform distribution $\bm{u}$ towards a one-hot distribution $\bm{o}$. Defined as:
$$ Loc = \frac{1-cos(\angle (\bm{p},\bm{u}))}{1-cos(\angle (\bm{o},\bm{u}))}$$
$Loc$ varies between 0 ($\bm{p} = \bm{u}$; zero certainty) and 1 ($\bm{p} = \bm{o}$; absolute certainty) and provides a good way to track the average knowledge distribution. Figure \ref{fig:loc} shows how this metric changes during training for the standard K-Mine model along with K-Mine-max and K-Mine-max-full. We can see that for K-Mine-max the localization stays close to zero (\textasciitilde 1e-4) whereas when knowledge labels are provided, it surges rapidly in less than 100 iterations. K-Mine shows a delayed (starting around 500 iterations) but more extreme surge. This can be attributed to the weighted fusion which forces the model to localize the distribution as soon as possible -and thus at the cost of a less accurate distribution- in order to pass a less noisy representation to the decoder. The model seems to partly re-evaluate and improve this process subsequently as can be seen in the slightly descending behaviour of $Loc$ in later steps, eventually converging to the ascending supervised values.

\section{Conclusion and future work}
In this work we introduced K-Mine (\underline{K}nowledge \underline{M}ixing \underline{in} \underline{e}ncoder), which provides a simple way to train knowledge grounded conversation models based on pretrained generative models without knowledge labels or a separate retriever. K-Mine uses a weighted aggregation method to fuse different encoded knowledge-context pairs into one sequence of the same length before passing it to the decoder, which has its advantages and disadvantages in comparison to models like RAG and FiD. It significantly reduces the computational cost in the decoder (at least by a factor of m= number of encoded knowledge options) but this naturally comes with the cost of a noisier and less rich input for the decoder which affects the response generation performance. This was partly confirmed by the relatively low KF1 values of K-Mine compared with the fully supervised version in Table \ref{wow-abalation} but a qualitative assessment of K-Mine's response generation can shed more light on this. 

Another interesting topic is the relationship between the KS and RG performance in K-Mine. In our experiments we saw that in later iterations, the KS performance often keeps improving while the RG performance (according to automatic metrics) starts to suffer. A more comprehensive study can determine whether it is possible to reconcile the two, or K-Mine actually serves better as an auxiliary retriever or re-ranker module. Either way, the potentials and limitations of this approach (i.e. selection via generation) in similar tasks and problems is worth exploring.   

\section{Acknowledgement}
This research received funding from the Flemish Government under the ``Onderzoeksprogramma Artificiële Intelligentie (AI) Vlaanderen'' program.

\bibliography{anthology,custom}

\begin{thebibliography}{24}
\expandafter\ifx\csname natexlab\endcsname\relax\def\natexlab#1{#1}\fi

\bibitem[{Bruyn et~al.(2020)Bruyn, Lotfi, Buhmann, and
  Daelemans}]{Bruyn2020BARTFK}
Maxime~De Bruyn, Ehsan Lotfi, Jeska Buhmann, and Walter Daelemans. 2020.
\newblock \href {http://ceur-ws.org/Vol-2666/KDD_Converse20_paper_7.pdf} {Bart
  for knowledge grounded conversations}.
\newblock In \emph{Converse@KDD}.

\bibitem[{Chen et~al.(2020)Chen, Meng, Li, Chen, Xu, Xu, and
  Zhou}]{chen-etal-2020-bridging}
Xiuyi Chen, Fandong Meng, Peng Li, Feilong Chen, Shuang Xu, Bo~Xu, and Jie
  Zhou. 2020.
\newblock \href {https://doi.org/10.18653/v1/2020.emnlp-main.275} {Bridging the
  gap between prior and posterior knowledge selection for knowledge-grounded
  dialogue generation}.
\newblock In \emph{Proceedings of the 2020 Conference on Empirical Methods in
  Natural Language Processing (EMNLP)}, pages 3426--3437, Online. Association
  for Computational Linguistics.

\bibitem[{Dinan et~al.(2019)Dinan, Roller, Shuster, Fan, Auli, and
  Weston}]{dinan2019wizard}
Emily Dinan, Stephen Roller, Kurt Shuster, Angela Fan, Michael Auli, and Jason
  Weston. 2019.
\newblock \href {http://arxiv.org/abs/1811.01241} {Wizard of wikipedia:
  Knowledge-powered conversational agents}.

\bibitem[{Ghazvininejad et~al.(2018)Ghazvininejad, Brockett, Chang, Dolan, Gao,
  tau Yih, and Galley}]{ghazvininejad2018knowledgegrounded}
Marjan Ghazvininejad, Chris Brockett, Ming-Wei Chang, Bill Dolan, Jianfeng Gao,
  Wen tau Yih, and Michel Galley. 2018.
\newblock \href {http://arxiv.org/abs/1702.01932} {A knowledge-grounded neural
  conversation model}.

\bibitem[{Gopalakrishnan et~al.(2019)Gopalakrishnan, Hedayatnia, Chen,
  Gottardi, Kwatra, Venkatesh, Gabriel, and Hakkani-Tür}]{Gopalakrishnan2019}
Karthik Gopalakrishnan, Behnam Hedayatnia, Qinlang Chen, Anna Gottardi, Sanjeev
  Kwatra, Anu Venkatesh, Raefer Gabriel, and Dilek Hakkani-Tür. 2019.
\newblock \href {https://doi.org/10.21437/Interspeech.2019-3079}
  {{Topical-Chat: Towards Knowledge-Grounded Open-Domain Conversations}}.
\newblock In \emph{Proc. Interspeech 2019}, pages 1891--1895.

\bibitem[{Humeau et~al.(2020)Humeau, Shuster, Lachaux, and
  Weston}]{humeau2020polyencoders}
Samuel Humeau, Kurt Shuster, Marie-Anne Lachaux, and Jason Weston. 2020.
\newblock \href {http://arxiv.org/abs/1905.01969} {Poly-encoders: Transformer
  architectures and pre-training strategies for fast and accurate
  multi-sentence scoring}.

\bibitem[{Izacard and Grave(2021)}]{izacard2021leveraging}
Gautier Izacard and Edouard Grave. 2021.
\newblock \href {http://arxiv.org/abs/2007.01282} {Leveraging passage retrieval
  with generative models for open domain question answering}.

\bibitem[{Kim et~al.(2020)Kim, Ahn, and Kim}]{kim2020sequential}
Byeongchang Kim, Jaewoo Ahn, and Gunhee Kim. 2020.
\newblock \href {http://arxiv.org/abs/2002.07510} {Sequential latent knowledge
  selection for knowledge-grounded dialogue}.

\bibitem[{Lewis et~al.(2019)Lewis, Liu, Goyal, Ghazvininejad, Mohamed, Levy,
  Stoyanov, and Zettlemoyer}]{lewis2019bart}
Mike Lewis, Yinhan Liu, Naman Goyal, Marjan Ghazvininejad, Abdelrahman Mohamed,
  Omer Levy, Ves Stoyanov, and Luke Zettlemoyer. 2019.
\newblock \href {http://arxiv.org/abs/1910.13461} {Bart: Denoising
  sequence-to-sequence pre-training for natural language generation,
  translation, and comprehension}.

\bibitem[{Lewis et~al.(2021)Lewis, Perez, Piktus, Petroni, Karpukhin, Goyal,
  Küttler, Lewis, tau Yih, Rocktäschel, Riedel, and
  Kiela}]{lewis2021retrievalaugmented}
Patrick Lewis, Ethan Perez, Aleksandra Piktus, Fabio Petroni, Vladimir
  Karpukhin, Naman Goyal, Heinrich Küttler, Mike Lewis, Wen tau Yih, Tim
  Rocktäschel, Sebastian Riedel, and Douwe Kiela. 2021.
\newblock \href {http://arxiv.org/abs/2005.11401} {Retrieval-augmented
  generation for knowledge-intensive nlp tasks}.

\bibitem[{Lian et~al.(2019)Lian, Xie, Wang, Peng, and Wu}]{lian2019learning}
Rongzhong Lian, Min Xie, Fan Wang, Jinhua Peng, and Hua Wu. 2019.
\newblock \href {http://arxiv.org/abs/1902.04911} {Learning to select knowledge
  for response generation in dialog systems}.

\bibitem[{Liu et~al.(2018)Liu, Chen, Ren, Feng, Liu, and
  Yin}]{liu-etal-2018-knowledge}
Shuman Liu, Hongshen Chen, Zhaochun Ren, Yang Feng, Qun Liu, and Dawei Yin.
  2018.
\newblock \href {https://doi.org/10.18653/v1/P18-1138} {Knowledge diffusion for
  neural dialogue generation}.
\newblock In \emph{Proceedings of the 56th Annual Meeting of the Association
  for Computational Linguistics (Volume 1: Long Papers)}, pages 1489--1498,
  Melbourne, Australia. Association for Computational Linguistics.

\bibitem[{Meng et~al.(2021)Meng, Ren, Chen, Ren, Xi, and
  Rijke}]{Meng2021InitiativeAwareSL}
Chuan Meng, Pengjie Ren, Zhumin Chen, Z.~Ren, Tengxiao Xi, and M.~Rijke. 2021.
\newblock Initiative-aware self-supervised learning for knowledge-grounded
  conversations.
\newblock \emph{Proceedings of the 44th International ACM SIGIR Conference on
  Research and Development in Information Retrieval}.

\bibitem[{Meng et~al.(2020)Meng, Ren, Chen, Sun, Ren, Tu, and
  Rijke}]{10.1145/3397271.3401097}
Chuan Meng, Pengjie Ren, Zhumin Chen, Weiwei Sun, Zhaochun Ren, Zhaopeng Tu,
  and Maarten~de Rijke. 2020.
\newblock \href {https://doi.org/10.1145/3397271.3401097} {Dukenet: A dual
  knowledge interaction network for knowledge-grounded conversation}.
\newblock In \emph{Proceedings of the 43rd International ACM SIGIR Conference
  on Research and Development in Information Retrieval}, SIGIR '20, page
  1151–1160, New York, NY, USA. Association for Computing Machinery.

\bibitem[{Moghe et~al.(2018)Moghe, Arora, Banerjee, and
  Khapra}]{moghe2018exploiting}
Nikita Moghe, Siddhartha Arora, Suman Banerjee, and Mitesh~M. Khapra. 2018.
\newblock \href {http://arxiv.org/abs/1809.08205} {Towards exploiting
  background knowledge for building conversation systems}.

\bibitem[{Radford et~al.(2019)Radford, Wu, Child, Luan, Amodei, and
  Sutskever}]{Radford2019LanguageMA}
Alec Radford, Jeff Wu, R.~Child, David Luan, Dario Amodei, and Ilya Sutskever.
  2019.
\newblock Language models are unsupervised multitask learners.

\bibitem[{Raffel et~al.(2020)Raffel, Shazeer, Roberts, Lee, Narang, Matena,
  Zhou, Li, and Liu}]{raffel2020exploring}
Colin Raffel, Noam Shazeer, Adam Roberts, Katherine Lee, Sharan Narang, Michael
  Matena, Yanqi Zhou, Wei Li, and Peter~J. Liu. 2020.
\newblock \href {http://arxiv.org/abs/1910.10683} {Exploring the limits of
  transfer learning with a unified text-to-text transformer}.

\bibitem[{Roller et~al.(2020)Roller, Boureau, Weston, Bordes, Dinan, Fan,
  Gunning, Ju, Li, Poff, Ringshia, Shuster, Smith, Szlam, Urbanek, and
  Williamson}]{roller2020opendomain}
Stephen Roller, Y-Lan Boureau, Jason Weston, Antoine Bordes, Emily Dinan,
  Angela Fan, David Gunning, Da~Ju, Margaret Li, Spencer Poff, Pratik Ringshia,
  Kurt Shuster, Eric~Michael Smith, Arthur Szlam, Jack Urbanek, and Mary
  Williamson. 2020.
\newblock \href {http://arxiv.org/abs/2006.12442} {Open-domain conversational
  agents: Current progress, open problems, and future directions}.

\bibitem[{Shuster et~al.(2021)Shuster, Poff, Chen, Kiela, and
  Weston}]{shuster2021retrieval}
Kurt Shuster, Spencer Poff, Moya Chen, Douwe Kiela, and Jason Weston. 2021.
\newblock \href {http://arxiv.org/abs/2104.07567} {Retrieval augmentation
  reduces hallucination in conversation}.

\bibitem[{Wolf et~al.(2020)Wolf, Debut, Sanh, Chaumond, Delangue, Moi, Cistac,
  Rault, Louf, Funtowicz, Davison, Shleifer, von Platen, Ma, Jernite, Plu, Xu,
  Scao, Gugger, Drame, Lhoest, and Rush}]{wolf2020huggingfaces}
Thomas Wolf, Lysandre Debut, Victor Sanh, Julien Chaumond, Clement Delangue,
  Anthony Moi, Pierric Cistac, Tim Rault, Rémi Louf, Morgan Funtowicz, Joe
  Davison, Sam Shleifer, Patrick von Platen, Clara Ma, Yacine Jernite, Julien
  Plu, Canwen Xu, Teven~Le Scao, Sylvain Gugger, Mariama Drame, Quentin Lhoest,
  and Alexander~M. Rush. 2020.
\newblock \href {http://arxiv.org/abs/1910.03771} {Huggingface's transformers:
  State-of-the-art natural language processing}.

\bibitem[{Zhan et~al.(2021)Zhan, Zhang, Chen, Ding, Bao, and
  Lan}]{zhan-etal-2021-augmenting}
Haolan Zhan, Hainan Zhang, Hongshen Chen, Zhuoye Ding, Yongjun Bao, and Yanyan
  Lan. 2021.
\newblock \href {https://doi.org/10.18653/v1/2021.naacl-main.446} {Augmenting
  knowledge-grounded conversations with sequential knowledge transition}.
\newblock In \emph{Proceedings of the 2021 Conference of the North American
  Chapter of the Association for Computational Linguistics: Human Language
  Technologies}, pages 5621--5630, Online. Association for Computational
  Linguistics.

\bibitem[{Zhao et~al.(2020)Zhao, Wu, Xu, Tao, Zhao, and
  Yan}]{zhao-etal-2020-knowledge}
Xueliang Zhao, Wei Wu, Can Xu, Chongyang Tao, Dongyan Zhao, and Rui Yan. 2020.
\newblock \href {https://doi.org/10.18653/v1/2020.emnlp-main.272}
  {Knowledge-grounded dialogue generation with pre-trained language models}.
\newblock In \emph{Proceedings of the 2020 Conference on Empirical Methods in
  Natural Language Processing (EMNLP)}, pages 3377--3390, Online. Association
  for Computational Linguistics.

\bibitem[{Zheng et~al.(2020)Zheng, Cao, Jiang, and
  Huang}]{zheng2020differenceaware}
Chujie Zheng, Yunbo Cao, Daxin Jiang, and Minlie Huang. 2020.
\newblock \href {http://arxiv.org/abs/2009.09378} {Difference-aware knowledge
  selection for knowledge-grounded conversation generation}.

\bibitem[{Zheng and Zhou(2019)}]{10.1145/3357384.3357889}
Wen Zheng and Ke~Zhou. 2019.
\newblock \href {https://doi.org/10.1145/3357384.3357889} {Enhancing
  conversational dialogue models with grounded knowledge}.
\newblock In \emph{Proceedings of the 28th ACM International Conference on
  Information and Knowledge Management}, CIKM '19, page 709–718, New York,
  NY, USA. Association for Computing Machinery.

\end{thebibliography}
\bibliographystyle{acl_natbib}

\end{document}